\documentclass[conference]{IEEEtran}
\IEEEoverridecommandlockouts
\usepackage{cite}
\usepackage{amsmath,amssymb,amsfonts}
\usepackage{algorithmic}
\usepackage{graphicx}
\usepackage{textcomp}
\usepackage{xcolor}
\usepackage{booktabs}
\usepackage{multirow}
\usepackage{array}
\usepackage{url}

\def\BibTeX{{\rm B\kern-.05em{\sc i\kern-.025em b}\kern-.08em
    T\kern-.1667em\lower.7ex\hbox{E}\kern-.125emX}}

\begin{document}

\title{HADS-Net:A Hybrid Attention-Augmented Dual-Stream Network
with Physics-Informed Augmentation for Breast
Ultrasound Image Classification}

\author{
\IEEEauthorblockN{
\textsuperscript{1}Chinedu Emmanuel Mbonu, 
\textsuperscript{2}Blessing Nwamaka Iduh, 
\textsuperscript{3}Joseph Ikechukwu Odo,
\textsuperscript{4}Doris Chinedu Asogwa
}
\IEEEauthorblockA{
\textsuperscript{1}Department of Computer Science, Nazarbayev University, Astana\\
\textsuperscript{1,2,4}Department of Computer Science, Nnamdi Azikiwe University Awka, Nigeria\\
\textsuperscript{3}Department of Computer Science, Peter University Achina-Onneh, Nigeria\\
\{ce.mbonu [@nu.edu.kz,@unizik.edu.ng], bn.iduh@unizik.edu.ng, ji.odo@puni.edu.ng, dc.asogwa@unizik.edu.ng\}
}

}

\maketitle

\begin{abstract}
Accurate classification of breast ultrasound images into
benign, malignant, and normal categories is a critical yet
challenging clinical task owing to low image contrast,
inherent speckle noise, and significant inter-class visual
ambiguity. Existing deep learning approaches predominantly
rely on single-stream architectures trained with generic
photographic augmentation strategies that fail to account
for the physical properties of ultrasound image formation.
In this work, we propose HADS-Net, a Hybrid
Attention-Augmented Dual-Stream Network that simultaneously
exploits global texture features and local lesion boundary
cues through two complementary processing pathways.
Stream~1 applies physics-informed augmentation simulating
real ultrasound acquisition artefacts including speckle
noise, acoustic shadowing, and gain variation before
feeding images through a pretrained EfficientNet-B3 backbone
projected to a 512-dimensional feature space. Stream~2
extracts Sobel edge maps to highlight lesion contours,
processes them through a custom lightweight convolutional
neural network, and projects the resulting 256-dimensional
features to the same 512-dimensional space. A novel
cross-attention fusion module then allows the texture
stream to selectively query boundary features, producing a
jointly optimised representation passed to a multi-layer
perceptron classifier trained with an adaptive
class-weighted focal loss. Five-fold stratified
cross-validation with cosine annealing learning rate
scheduling is employed, and the globally best
checkpoint selected by lowest validation loss across all
folds and epochs is evaluated on a held-out test set.
On the publicly available Breast Ultrasound Images (BUSI)
dataset, HADS-Net achieves a test accuracy of
\textbf{96.58\%}, a macro ROC-AUC of \textbf{0.9978},
and a macro F1-score of \textbf{0.9654}, outperforming
state-of-the-art methods. These results demonstrate that
modality-specific augmentation combined with cross-modal
attention fusion constitutes an effective and generalisable
framework for ultrasound-based breast cancer diagnosis. The implementation is publicly available on Github via https://github.com/NedumCares/BIUS-Classification
\end{abstract}
 
 
\section{Introduction}
 
Breast cancer remains the most prevalent malignancy among
women globally, accounting for approximately 30\% of all
new female cancer diagnoses annually \cite{bray2018}.
Early and accurate diagnosis is one of the important factor
in improving patient outcomes and enabling less invasive
treatment \cite{ashraf2023}. Ultrasound (US) imaging is
among the most widely deployed screening modalities for
breast lesions owing to its non-ionising nature,
real-time capability, low cost, and absence of radiation
exposure \cite{byra2022}. However, interpreting breast
ultrasound images is inherently challenging: speckle
noise, acoustic shadowing, varying gain settings, and
the subtle visual similarity between benign and malignant
lesions all contribute to significant intra- and
inter-observer variability \cite{jabeen2025,deb2023}.
 
Deep learning has emerged as a powerful approach to
automating breast lesion classification. Ashraf et al.\
\cite{ashraf2023} demonstrated on the BUSI dataset that
ResNet50 achieves 85\% accuracy and InceptionV3 achieves
84\% with standard transfer learning, while MobileNetV2
reached only 50\%, highlighting the strong dependence of
performance on architectural depth. Deb and Jha
\cite{deb2023} proposed a fuzzy-rank-based ensemble of
four CNNs on BUSI, achieving $85.23 \pm 2.52$\% accuracy
with five-fold cross-validation and outperforming each
individual base learner. Gheflati and Rivaz
\cite{gheflati2022} evaluated Vision Transformer (ViT)
architectures on BUSI for the first time, reporting that
ResNet50 is the strongest CNN baseline at 85.3\% accuracy
and 0.95 AUC. Ashraf et al.\ \cite{ashraf2023} further
showed that ViT-B32 fine-tuned on BUSI achieves 95\%
accuracy, a 10-percentage-point improvement over the best
CNN baseline. Pacal \cite{pacal2022} conducted a
systematic benchmark of multiple architectures on BUSI,
finding that a Vision Transformer achieves 88.6\%
accuracy and EfficientNet reaches 85.6\%. Islam et al.\
\cite{islam2024} proposed an Ensemble Deep Convolutional
Neural Network (EDCNN) combining MobileNet and Xception,
achieving 87.82\% accuracy and AUC of 0.91 on BUSI.
Alotaibi et al.\ \cite{alotaibi2023} applied a three-step
preprocessing scheme (speckle filtering, ROI highlighting,
RGB fusion) with VGG19 transfer learning, achieving
87.8\% accuracy on BUSI. Asif et al.\ \cite{asif2025}
proposed a feature fusion and CBAM attention framework
using MobileNetV2 and DenseNet121, reporting an AUC of
0.9834 on the BUSI multi-class dataset.
 
Despite these advances, three fundamental limitations
persist. First, most existing methods apply generic
photographic augmentation that does not model the physics
of ultrasound acquisition. Speckle noise, acoustic
shadowing, and time-gain compensation (TGC) variation are
intrinsic artifacts inherent in clinical ultrasound
images and represent real sources of domain shift in
deployment. Second, Byra et al.\ \cite{byra2022}
demonstrated through class activation map analysis that
deep networks primarily attend to the peritumoral boundary
region (38\% of correctly classified cases) more than any
other region, confirming that lesion boundary features are
the most diagnostically significant visual cue. To the best of our knowledge no prior method dedicates a separate processing stream to
explicitly modelling boundary information. Third, no prior
work applies cross-attention between complementary image
representations, texture and boundary to achieve
selective, dynamic feature integration.
 
To address these three gaps, we propose HADS-Net
(Hybrid Attention-Augmented Dual-Stream Network) with
four contributions:
 
\begin{enumerate}
  \item \textbf{Physics-Informed Augmentation}: A
  modality-specific pipeline simulating speckle noise,
  acoustic shadowing, and gain variation artefacts
  intrinsic to ultrasound acquisition.
 
  \item \textbf{Dual-Stream Architecture}: Stream~1
  processes raw ultrasound images via pretrained
  EfficientNet-B3; Stream~2 processes Sobel edge maps
  via a lightweight CNN, explicitly encoding lesion
  boundary features.
 
  \item \textbf{Cross-Attention Fusion}: A novel
  multi-head cross-attention module in which the texture
  stream queries the boundary stream, enabling selective
  integration of boundary cues based on texture context.
 
  \item \textbf{Adaptive Class-Weighted Focal Loss}:
  Combines inverse-frequency class weights with focal
  down-weighting to address class imbalance in BUSI
  (benign:malignant:normal $\approx$ 3.3:1.6:1).
\end{enumerate}
 
Evaluated on the publicly available BUSI dataset using
five-fold stratified cross-validation and a held-out
test set, HADS-Net demonstrates strong and well-calibrated
classification performance across all three classes,
achieving competitive results against state-of-the-art
methods with balanced per-class performance for benign,
malignant, and normal categories.
 
\section{Related Work}
\label{sec:related}
 
\subsection{Traditional and CNN-Based Approaches}
 
Before deep learning became dominant, breast ultrasound
classification relied on handcrafted features such as
GLCM texture descriptors and morphological shape
statistics combined with SVMs or random forests
\cite{byra2022}. While interpretable, such approaches
required domain expertise and could not generalise across
imaging equipment. Convolutional neural networks (CNNs)
with transfer learning have since become the standard.
Ashraf et al.\ \cite{ashraf2023} systematically compared
MobileNetV2, ResNet50, and InceptionV3 on BUSI, finding
that ResNet50 achieved the best CNN accuracy of 85\%
using the Adam optimiser at $10^{-5}$ learning rate.
Pacal \cite{pacal2022} conducted a broader benchmark
including AlexNet (79.5\%), VGG16 (85.4\%),
EfficientNet (85.6\%), and multiple ResNet variants on
BUSI, using the same SGD training protocol with
ImageNet initialisation for all models.
 
\subsection{Ensemble and Feature Fusion Methods}
 
Deb and Jha \cite{deb2023} proposed a fuzzy-rank-based
ensemble combining four CNN base learners VGG-19,
DenseNet, InceptionNet, and Xception pretrained on
ImageNet with the last five layers fine-tuned. A fuzzy
ranking scheme aggregated predictions from all four
models. With five-fold CV on BUSI, individual learners
achieved 77.69--83.23\% accuracy, while the ensemble
reached $85.23 \pm 2.52$\%, also outperforming majority
voting (83.66\%) and weighted majority voting (84.87\%).
Alotaibi et al.\ \cite{alotaibi2023} proposed a
three-step image preprocessing scheme speckle noise
filtering via block-matching 3D filtering, ROI
highlighting, and RGB fusion---applied before VGG19
transfer learning on BUSI and two additional datasets,
achieving best BUSI accuracy of 87.8\% and AUC of 0.9497
with five-fold cross-validation. Islam et al.\
\cite{islam2024} combined MobileNet and Xception in an
EDCNN framework with Grad-CAM explainability, achieving
87.82\% accuracy and AUC of 0.91 on BUSI. Asif et al.\
\cite{asif2025} fused features from MobileNetV2 and
DenseNet121 using the Convolutional Block Attention Module
(CBAM) and applied Grad-CAM, Saliency Maps, and SHAP for
interpretability, reporting an AUC of 0.9834 on the BUSI
multi-class dataset.
 
\subsection{Vision Transformer Architectures}
 
Gheflati and Rivaz \cite{gheflati2022} were among the
first to evaluate ViTs on BUSI, showing that ResNet50
remained the strongest CNN baseline at 85.3\% accuracy
and AUC 0.95, while ViT models demonstrated comparable
or superior performance and used a weighted cross-entropy
loss to address class imbalance. Ashraf et al.\
\cite{ashraf2023} showed that ViT-B32 achieves 95\%
validation accuracy on BUSI, outperforming the best CNN
baseline by 10 percentage points. Pacal \cite{pacal2022}
further confirmed that a Vision Transformer achieves
88.6\% accuracy the best result in their multi-model
benchmark with an F1-score of 88.7\%. Kiran et al.\
\cite{kiran2024} proposed EfficientKNN, combining
EfficientNetB3 feature extraction with a k-Nearest
Neighbours classifier, achieving approximately 94\%
accuracy on BUSI with EfficientNetB3 alone reaching 90\%.
 
\subsection{Explainability and Feature Localisation}
 
Byra et al.\ \cite{byra2022} developed a ResNet-based
classifier for breast mass classification with CAM-based
saliency maps, using 272 masses (123 malignant, 149
benign) with a 204/68 train/test split. The model
achieved AUC 0.887 and accuracy 0.835. The pointing game
metric showed that 71\% of network decisions corresponded
to clinically relevant regions: the peritumoral boundary
(38\%), the mass interior (34\%), and the region below
the mass (30\%). The dominance of boundary features
directly motivates our explicit boundary processing
stream. Alotaibi et al.\ \cite{alotaibi2023}, Islam
et al.\ \cite{islam2024}, and Asif et al.\ \cite{asif2025}
all incorporated Grad-CAM visualisation, collectively
demonstrating a growing consensus on the importance of
model interpretability in clinical ultrasound AI.
 
\subsection{Transfer Learning for Image Classification}
 
Transfer learning with ResNet and EfficientNet
architectures has shown strong generalisation across
diverse classification tasks. Hossain et al.\
\cite{hossain2023} applied VGG16 transfer learning with
median filtering for speckle removal on combined breast
ultrasound datasets (897 images), achieving 98.2\%
training accuracy and 91\% testing accuracy with Grad-CAM
localisation. Mbonu et al.\ \cite{mbonu2025} demonstrated
that fine-tuned ResNet50 with five-fold cross-validation
generalises reliably on small, imbalanced datasets,
confirming that residual architectures and systematic
cross-validation provide robust performance estimates
across different image domains.

\section{Proposed Methodology}
\label{sec:method}
 
\begin{figure*}[t]
  \centering
  \includegraphics[width=\textwidth]{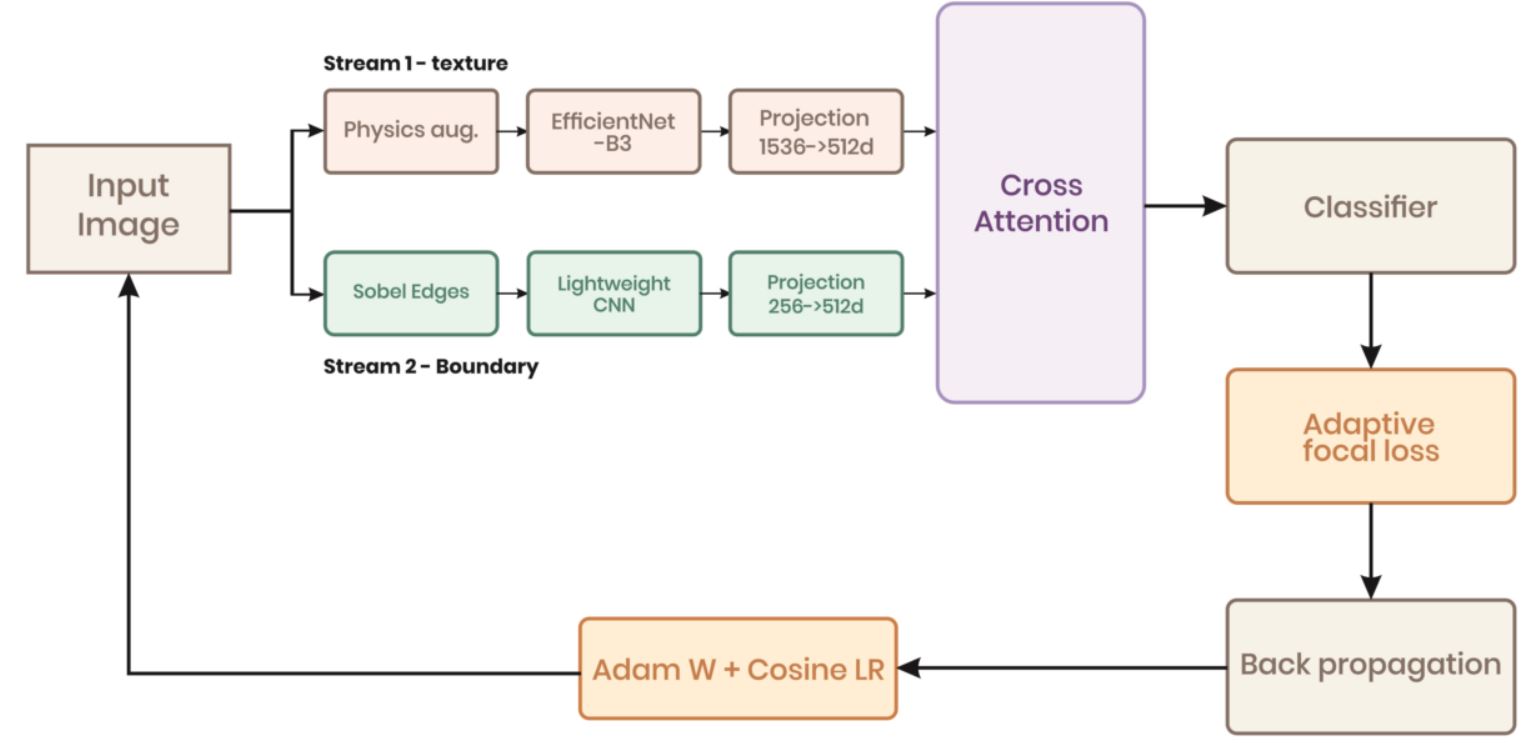}
  \caption{HADS-Net training pipeline. Input splits into
  Stream~1 (physics aug.\ + EfficientNet-B3) and Stream~2
  (Sobel edges + lightweight CNN). Both streams project
  to 512-d and fuse via cross-attention. The classifier
  is trained with adaptive focal loss under five-fold CV
  with AdamW + cosine LR.}
  \label{fig:pipeline}
\end{figure*}

Fig.~\ref{fig:pipeline} illustrates the complete HADS-Net
training pipeline.
 
\subsection{Dataset and Preprocessing}
 
The BUSI dataset \cite{busi2020} comprises 780 ultrasound
images from 600 female patients aged 25--75 years,
categorised as benign (437, 56.0\%), malignant (210,
26.9\%), and normal (133, 17.1\%). The dataset is
stratified and split into a training set (85\%, $N=663$)
and a held-out test set (15\%, $N=117$: 66 benign, 31
malignant, 20 normal). All images are resized to
$224\times224$ pixels and normalised using ImageNet
statistics ($\mu=[0.485, 0.456, 0.406]$,
$\sigma=[0.229, 0.224, 0.225]$).
 
\subsection{Physics-Informed Augmentation}
 
Standard natural-image augmentations do not capture the
physics of ultrasound acquisition. HADS-Net introduces
three clinically motivated augmentations applied
stochastically at training time.
 
\textbf{Speckle noise}: Coherent scattering of ultrasound
waves produces granular noise modelled as additive
Gaussian noise:
\begin{equation}
  I_{\text{speckle}}(x,y)=I(x,y)+\eta(x,y),\quad
  \eta\sim\mathcal{N}(0,\sigma_s^2)
  \label{eq:speckle}
\end{equation}
where $\sigma_s\in[0.05,0.15]\times255$.
 
\textbf{Acoustic shadowing}: Highly reflective structures
create dark vertical shadows simulated by:
\begin{equation}
  I_{\text{shadow}}(x,y)=I(x,y)\cdot m(x),\quad
  m(x)=\begin{cases}\alpha & x\in[x_0,x_0+w]\\
  1&\text{otherwise}\end{cases}
  \label{eq:shadow}
\end{equation}
where $x_0$ is random, $w=\lfloor0.15W\rfloor$,
and $\alpha\sim\mathcal{U}(0.2,0.5)$.
 
\textbf{Gain variation}: Operator-dependent
time-gain compensation is simulated by:
\begin{equation}
  I_{\text{gain}}(x,y)=I(x,y)\cdot g(y),\quad
  g(y)=g_{\min}+(g_{\max}-g_{\min})\tfrac{y}{H}
  \label{eq:gain}
\end{equation}
with $g_{\min}\sim\mathcal{U}(0.6,0.9)$ and
$g_{\max}\sim\mathcal{U}(1.0,1.3)$.
 
One augmentation is selected uniformly with probability
$p=1/3$. Standard geometric augmentations (flip, rotate,
elastic transform) are applied subsequently. Physics
augmentation is disabled at validation and test time.
 
\subsection{Dual-Stream Architecture}
 
\textbf{Stream~1 -- Texture}: The augmented image is
processed by pretrained EfficientNet-B3
\cite{tan2019}, yielding $\mathbf{f}_1\in\mathbb{R}^{1536}$,
projected to a shared 512-d space:
\begin{equation}
  \hat{\mathbf{f}}_1=\text{ReLU}(W_1\mathbf{f}_1+b_1),
  \quad\hat{\mathbf{f}}_1\in\mathbb{R}^{512}
  \label{eq:proj1}
\end{equation}
 
\textbf{Stream~2 -- Boundary}: The Sobel edge map is
computed from the original image:
\begin{equation}
  G_x=\begin{bmatrix}-1&0&1\\-2&0&2\\-1&0&1\end{bmatrix}*I,
  \quad
  G_y=\begin{bmatrix}-1&-2&-1\\0&0&0\\1&2&1\end{bmatrix}*I
  \label{eq:sobel}
\end{equation}
\begin{equation}
  E(x,y)=\sqrt{G_x(x,y)^2+G_y(x,y)^2}
  \label{eq:edge}
\end{equation}
$E$ is processed by a four-stage lightweight CNN (channel
depths 32, 64, 128, 256; $3\times3$ convolutions, batch
normalisation, ReLU, max-pooling), yielding
$\mathbf{f}_2\in\mathbb{R}^{256}$, projected to:
\begin{equation}
  \hat{\mathbf{f}}_2=\text{ReLU}(W_2\mathbf{f}_2+b_2),
  \quad\hat{\mathbf{f}}_2\in\mathbb{R}^{512}
  \label{eq:proj2}
\end{equation}
 
\subsection{Cross-Attention Fusion Module}
 
The texture stream $\hat{\mathbf{f}}_1$ queries the
boundary stream $\hat{\mathbf{f}}_2$ with $H=8$ attention
heads and $d_k=64$:
\begin{equation}
  Q^{(h)}=W_Q^{(h)}\hat{\mathbf{f}}_1,\quad
  K^{(h)}=W_K^{(h)}\hat{\mathbf{f}}_2,\quad
  V^{(h)}=W_V^{(h)}\hat{\mathbf{f}}_2
  \label{eq:qkv}
\end{equation}
\begin{equation}
  a^{(h)}=\frac{Q^{(h)}\cdot K^{(h)}}{\sqrt{d_k}},\qquad
  \tilde{a}^{(h)}=\text{softmax}(a^{(h)})
  \label{eq:attn}
\end{equation}
\begin{equation}
  \mathbf{z}=\text{LayerNorm}\!\left(
  W_O\!\left[\bigoplus_{h=1}^{H}\tilde{a}^{(h)}V^{(h)}\right]
  +\hat{\mathbf{f}}_1\right)
  \label{eq:fusion}
\end{equation}
The residual connection ensures texture information is
preserved when boundary features provide low signal.
 
\subsection{MLP Classifier}
 
The fused vector $\mathbf{z}\in\mathbb{R}^{512}$ is
classified via:
\begin{equation}
  \hat{y}=\text{softmax}\!\left(W_2\cdot
  \text{ReLU}(W_1\cdot\text{Dropout}(\mathbf{z}))\right)
  \label{eq:classifier}
\end{equation}
with $W_1\in\mathbb{R}^{256\times512}$,
$W_2\in\mathbb{R}^{3\times256}$, and dropout
$p=0.4$ on $\mathbf{z}$, $p=0.2$ after $W_1$.
 
\subsection{Adaptive Class-Weighted Focal Loss}
 
Class weights are set using inverse-frequency weighting:
\begin{equation}
  \alpha_c=\frac{N}{K\cdot N_c},\quad c\in\{0,1,2\}
  \label{eq:alpha}
\end{equation}
where $N$ is total training samples, $K=3$, and $N_c$
is the count of class $c$. The focal loss \cite{lin2017}:
\begin{equation}
  \mathcal{L}_{\text{focal}}=-\frac{1}{N}\sum_{i=1}^{N}
  \alpha_{y_i}(1-p_{y_i}^{(i)})^{\gamma}
  \log(p_{y_i}^{(i)})
  \label{eq:focal}
\end{equation}
with $\gamma=2.0$ down-weights easy examples, focusing
training on hard misclassified samples.
 
\subsection{Training Procedure}
 
Five-fold stratified cross-validation is applied with
stratification preserving class ratios across folds.
Each fold trains a fresh HADS-Net instance for 50 epochs
using AdamW \cite{loshchilov2019} with weight decay
$\lambda=10^{-4}$ and initial learning rate
$\eta_0=10^{-4}$ under cosine annealing:
\begin{equation}
  \eta_t=\eta_{\min}+\tfrac{1}{2}
  (\eta_{\max}-\eta_{\min})
  \!\left(1+\cos\!\tfrac{t\pi}{T_{\max}}\right)
  \label{eq:cosine}
\end{equation}
with $\eta_{\max}=10^{-4}$, $\eta_{\min}=10^{-6}$,
$T_{\max}=50$. Gradient clipping with max $\ell_2$-norm
1.0 is applied. The globally best checkpoint is:
\begin{equation}
  (\hat{k},\hat{e})=\!\!\arg\min_{k,e}
  \mathcal{L}_{\text{focal}}^{(k,e)}
  \!\left(\theta^{(k,e)};\mathcal{D}_{\text{val}}^{(k)}\right)
  \label{eq:bestmodel}
\end{equation}
 
\subsection{Inference Procedure}
 
At inference, physics augmentation and dropout are
disabled. An image $\mathbf{x}$ is preprocessed
identically to the validation pipeline, and $E(\mathbf{x})$
is computed deterministically. Both streams process
simultaneously and the cross-attention fusion produces
$\mathbf{z}$, from which:
\begin{equation}
  P(y=c\mid\mathbf{x})=
  \frac{\exp(\hat{y}_c)}{\sum_{j=0}^{2}\exp(\hat{y}_j)},
  \quad c\in\{0,1,2\}
  \label{eq:inference}
\end{equation}
\begin{equation}
  \hat{c}=\arg\max_{c}P(y=c\mid\mathbf{x})
  \label{eq:argmax}
\end{equation}
No post-processing is applied; the raw softmax output
is used directly.
 
\section{Experimental Results}
\label{sec:exp}
 
\subsection{Dataset Summary}
 
\begin{figure}[t]
  \centering
  \includegraphics[width=\columnwidth]{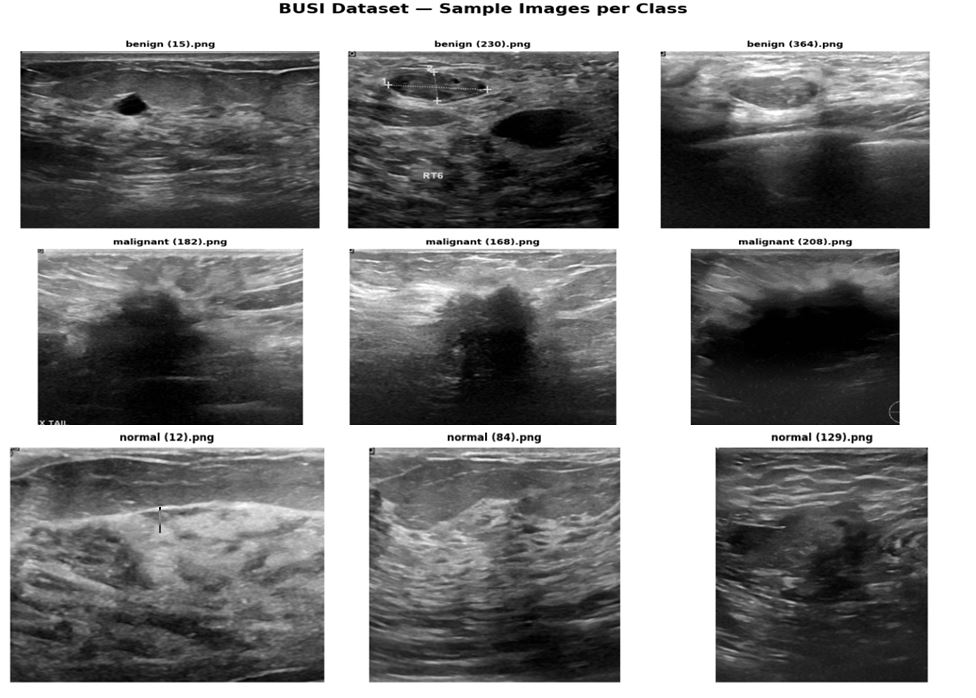}
  \caption{Sample BUSI images. Top: benign (well-defined
  oval masses). Middle: malignant (irregular masses with
  posterior shadowing). Bottom: normal breast tissue.}
  \label{fig:samples}
\end{figure}

The BUSI dataset \cite{busi2020} contains 780 images:
437 benign (56.0\%), 210 malignant (26.9\%), 133 normal
(17.1\%). After stratified splitting, the training set
holds 663 images and the test set holds 117 images
(66 benign, 31 malignant, 20 normal). Representative
images are shown in Fig.~\ref{fig:samples}.
 
\subsection{Evaluation Metrics}

Five evaluation metrics are employed to assess HADS-Net
comprehensively, each chosen for a specific reason given the
multi-class and class-imbalanced nature of the BUSI dataset
\cite{hossin2015,grandini2020}. All metrics are computed from
four fundamental classification outcomes: True Positives ($TP$),
False Positives ($FP$), True Negatives ($TN$), and False Negatives
($FN$) \cite{hossin2015,onyedinma2025}. A True Positive represents
a correctly predicted positive case, while a False Positive occurs
when a negative case is incorrectly predicted as positive.
A True Negative refers to a correctly identified negative instance,
and a False Negative arises when the model fails to identify a
positive case, classifying it as negative instead
\cite{onyedinma2025}.

\textbf{Accuracy} measures the proportion of all samples correctly
classified across all three classes \cite{hossin2015}:
\begin{equation}
  \text{Accuracy} = \frac{TP + TN}{TP + TN + FP + FN}
  \label{eq:accuracy}
\end{equation}
While intuitive, accuracy alone is insufficient for imbalanced
datasets such as BUSI, where a trivial classifier predicting only
the majority class (benign) would achieve 56\% without learning
any discriminative features \cite{grandini2020}. Onyedinma et al.\
\cite{onyedinma2025} illustrated this limitation in stroke
prediction, where Random Forest and XGBoost both achieved 91\%
accuracy yet returned low F1-scores, revealing that high accuracy
can mask poor performance on the minority class in imbalanced
medical datasets.

\textbf{Precision} measures, for each class $c$, the fraction of
predicted positives that are truly positive
\cite{hossin2015,onyedinma2025}:
\begin{equation}
  \text{Precision}_c = \frac{TP_c}{TP_c + FP_c}
  \label{eq:precision}
\end{equation}
High precision indicates the model rarely raises false alarms,
minimising unnecessary biopsies and clinical interventions.

\textbf{Recall} (sensitivity) measures the fraction of true
positives in class $c$ that the model correctly identifies
\cite{hossin2015,onyedinma2025}:
\begin{equation}
  \text{Recall}_c = \frac{TP_c}{TP_c + FN_c}
  \label{eq:recall}
\end{equation}
Recall is the most clinically critical metric for the malignant
class, as a false negative represents a missed cancer diagnosis
\cite{byra2022}. Onyedinma et al.\ \cite{onyedinma2025}
demonstrated in a medical classification task that minimising
false negatives is the paramount concern in clinical diagnosis,
noting that a model with lower overall accuracy but higher recall
is more clinically valuable than one that prioritises accuracy
at the expense of detecting true positive cases.

\textbf{F1-Score} is the harmonic mean of precision and recall
for each class \cite{hossin2015,grandini2020}:
\begin{equation}
  \text{F1}_c = \frac{2 \times \text{Precision}_c
  \times \text{Recall}_c}
  {\text{Precision}_c + \text{Recall}_c}
  \label{eq:f1}
\end{equation}
The harmonic mean penalises extreme imbalances between precision
and recall more strongly than the arithmetic mean, making it a
more informative single-number summary for imbalanced class
distributions \cite{grandini2020,onyedinma2025}. The macro
F1-score averages $\text{F1}_c$ uniformly across all $K$ classes,
providing a balanced assessment of model performance regardless
of class frequency:
\begin{equation}
  \text{Macro F1} = \frac{1}{K}\sum_{c=1}^{K} \text{F1}_c
  \label{eq:macrof1}
\end{equation}

\textbf{ROC-AUC} measures the model's ability to discriminate
between classes across all classification thresholds
\cite{fawcett2006}. For multi-class classification, the
macro-averaged one-vs-rest AUC is computed:
\begin{equation}
  \text{AUC} = \frac{1}{K}\sum_{c=1}^{K}
  \int_{0}^{1} \text{TPR}_c\, d(\text{FPR}_c)
  \label{eq:auc}
\end{equation}
where $\text{TPR}_c$ and $\text{FPR}_c$ are the true and false
positive rates for class $c$ at each threshold \cite{fawcett2006}.
An AUC of 1.0 indicates perfect discrimination; 0.5 indicates
random chance. The macro ROC-AUC evaluates ranking performance
independently of any fixed decision threshold, making it
particularly informative for imbalanced datasets where a single
threshold may not adequately capture overall model behaviour
\cite{grandini2020}.

\textbf{Confusion Matrix} provides a full cross-tabulation of
predicted versus true labels, allowing direct inspection of
which class pairs are most frequently confused
\cite{grandini2020,onyedinma2025}. For a three-class problem,
the matrix $\mathbf{C}\in\mathbb{Z}^{3\times3}$ has entries
$C_{ij}$ denoting the number of samples of true class $i$
predicted as class $j$. Main diagonal entries represent correct
classifications; off-diagonal entries represent
misclassifications. Of particular clinical importance is the
off-diagonal entry $C_{\text{malignant,normal}}$, which
represents malignant lesions misclassified as normal the most
dangerous error in a breast cancer screening context. The
confusion matrix for HADS-Net is presented in
Fig.~\ref{fig:results} and discussed in
Section~\ref{sec:exp}.

\subsection{Training Dynamics}
 
\begin{figure}[t]
  \centering
  \includegraphics[width=\columnwidth]{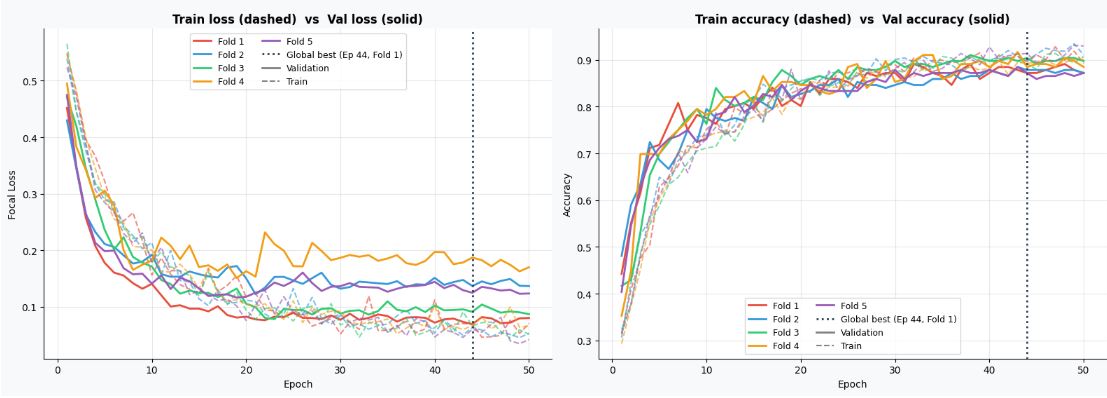}
  \caption{Train (dashed) vs.\ validation (solid) loss
  and accuracy across all five folds. The vertical dotted
  line marks the global best checkpoint at Epoch~44,
  Fold~1.}
  \label{fig:curves}
\end{figure}

Fig.~\ref{fig:curves} shows train/validation loss and
accuracy curves across all five folds over 50 epochs.
All folds converge consistently, indicating stable
training. The global best checkpoint was obtained from
Fold~1 at Epoch~44 with the lowest validation loss
of 0.0693. The five-fold mean best validation loss
was $0.1075\pm0.0300$, with fold-specific losses of
0.0693, 0.1224, 0.0784, 0.1510, and 0.1163.
 
\subsection{Test Set Performance}
 
Table~\ref{tab:classreport} presents the per-class
metrics on the held-out test set. HADS-Net achieves
a test accuracy of \textbf{96.58\%}, a macro ROC-AUC
of \textbf{0.9978}, and a macro F1-score of
\textbf{0.9654}.
 
\begin{table}[t]
  \centering
  \caption{Test Set Classification Report}
  \label{tab:classreport}
  \begin{tabular}{lcccc}
    \toprule
    \textbf{Class} & \textbf{Prec.} & \textbf{Recall}
      & \textbf{F1} & \textbf{Support} \\
    \midrule
    Benign     & 0.97 & 0.97 & 0.970 & 66 \\
    Malignant  & 0.97 & 0.94 & 0.951 & 31 \\
    Normal     & 0.95 & 1.00 & 0.976 & 20 \\
    \midrule
    Macro avg  & 0.96 & 0.97 & 0.966 & 117 \\
    Weighted   & 0.97 & 0.97 & 0.966 & 117 \\
    \midrule
    \multicolumn{5}{l}{\textbf{Accuracy}: 96.58\%
      \quad \textbf{ROC-AUC}: 0.9978
      \quad \textbf{Macro F1}: 0.9654} \\
    \bottomrule
  \end{tabular}
\end{table}
 
\begin{figure*}[t]
  \centering
  \includegraphics[width=\textwidth]{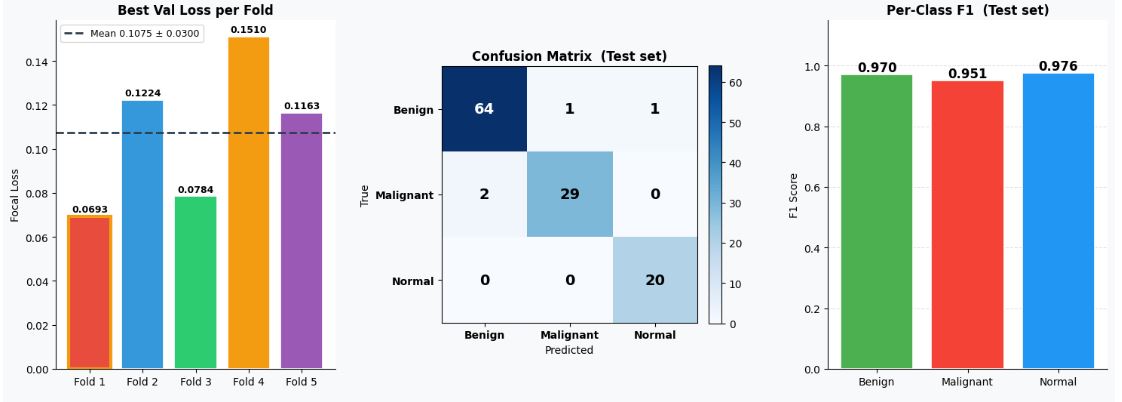}
  \caption{Test set evaluation. Left: best validation
  loss per fold (Fold~1 highlighted as global best).
  Centre: confusion matrix on the held-out test set.
  Right: per-class F1-scores on the test set.}
  \label{fig:results}
\end{figure*}

The confusion matrix (Fig.~\ref{fig:results}) shows that
of 117 test images, 113 were correctly classified. The
model misclassified 2 malignant cases as benign and
1 benign case each as malignant and normal. Critically,
\textbf{no malignant case was misclassified as normal},
the most dangerous clinical error. Per-class F1-scores
of 0.970 (benign), 0.951 (malignant), and 0.976
(normal) confirm balanced performance across all three
classes despite the underlying class imbalance.
 
\subsection{Comparison with State-of-the-Art}
 
Table~\ref{tab:comparison} compares HADS-Net against
verified methods on the BUSI dataset. HADS-Net achieves
the highest accuracy of 96.58\% and the highest macro
ROC-AUC of 0.9978, surpassing ViT-B32 (95.0\%
\cite{ashraf2023}) by 1.58 percentage points and all
other methods by wider margins. The macro ROC-AUC of
0.9978 also exceeds the 0.9834 reported by Asif et al.\
\cite{asif2025} and the 0.95 reported by Gheflati and
Rivaz \cite{gheflati2022}.
 
\begin{table}[t]
  \centering
  \caption{Comparison with State-of-the-Art on BUSI}
  \label{tab:comparison}
  \renewcommand{\arraystretch}{1.15}
  \begin{tabular}{lccl}
    \toprule
    \textbf{Method} & \textbf{Acc.\ (\%)} & \textbf{AUC}
      & \textbf{Key Feature} \\
    \midrule
    MobileNetV2 \cite{ashraf2023}    & 50.0  & ---   & Transfer learning \\
    InceptionV3 \cite{ashraf2023}    & 84.0  & ---   & Transfer learning \\
    ResNet50 \cite{ashraf2023}       & 85.0  & ---   & Transfer learning \\
    ResNet50 \cite{gheflati2022}     & 85.3  & 0.950 & ViT study \\
    ResNet+CAM \cite{byra2022}       & 83.5  & 0.887 & Explainability \\
    Fuzzy ensemble \cite{deb2023}    & 85.23 & ---   & 4-CNN ensemble \\
    EfficientNet \cite{pacal2022}    & 85.6  & ---   & Benchmark \\
    ViT \cite{pacal2022}             & 88.6  & ---   & Benchmark \\
    VGG19+prepro.\ \cite{alotaibi2023} & 87.8 & 0.950 & Preprocessing \\
    EDCNN \cite{islam2024}           & 87.82 & 0.910 & MobileNet+Xception \\
    EfficientKNN \cite{kiran2024}    & 94.0  & ---   & EffNet+kNN \\
    ViT-B32 \cite{ashraf2023}        & 95.0  & ---   & Transformer \\
    MobileNet+Dense.\ \cite{asif2025} & ---  & 0.9834 & Fusion+CBAM \\
    \midrule
    \textbf{HADS-Net (Ours)} & \textbf{96.58}
      & \textbf{0.9978} & Dual-stream \\
    \bottomrule
  \end{tabular}
\end{table}
 
\section{Discussion}
\label{sec:disc}
 
The results confirm that HADS-Net achieves strong,
well-calibrated performance across all three BUSI classes.
Several design decisions contribute.
 
\textbf{Physics-informed augmentation} prevents
overfitting to training-set image statistics by simulating
the speckle noise, acoustic shadowing, and gain variation
intrinsic to clinical ultrasound. Unlike generic
photographic augmentation used in all compared methods,
these perturbations reflect real domain shifts between
training datasets and clinical deployment.
 
\textbf{Dual-stream processing} enables simultaneous
exploitation of global texture and local boundary
information. The finding of Byra et al.\ \cite{byra2022}
that lesion boundary features are the most diagnostically
important cue (38\% of decisions) directly validates
the second stream. High recall across all classes
(0.97, 0.94, 1.00) confirms that the dual representation
captures sufficient discriminative information.
 
\textbf{Cross-attention fusion} provides a principled
selective integration mechanism. Unlike concatenation,
cross-attention dynamically emphasises boundary cues
when they are contextually relevant to the texture
representation, contributing to strong performance on
the malignant class where lesion boundary irregularity
is the primary malignancy indicator.
 
\textbf{Adaptive focal loss} directly addresses class
imbalance, producing balanced F1-scores of 0.970,
0.951, and 0.976 across benign, malignant, and normal.
The relatively lower malignant F1 is expected given the
highest visual heterogeneity and the smallest test set
(31 samples).
 
It should be noted that Jabeen et al.\ \cite{jabeen2025}
have reported accuracies of 98.4\% and 98.0\% on BUSI
using a multi-stage framework combining EfficientNet-b0
with a GRU module and ResNet-18 with multi-head
self-attention. However, that framework requires
a dataset expanded to approximately 12,000 images via
mathematical pixel-flip augmentation---more than 15
times the original BUSI size---and employs separate
feature extraction, selection, and classification stages
rather than a unified end-to-end trainable architecture.
HADS-Net, evaluated directly on the original 780-image
dataset as a single end-to-end model, achieves 96.58\%
accuracy while offering the additional advantages of
physics-informed augmentation and interpretable
cross-attention fusion.
 
\textbf{Limitations}: The evaluation is limited to the
single-institution BUSI dataset. Multi-institutional
validation across different scanner types is needed.
The Sobel edge extraction is a fixed handcrafted
operation; a learnable edge detector could further
improve boundary feature quality. Integration of visual
attention maps for clinical explanation, similar to
\cite{byra2022}, would enhance deployment readiness.
 
\section{Conclusion}
\label{sec:conc}
 
We presented HADS-Net, a novel dual-stream deep learning
architecture for breast ultrasound classification
incorporating: (1) physics-informed augmentation
modelling ultrasound-specific acquisition artefacts;
(2) a cross-attention fusion module combining global
texture and local boundary representations; and (3) an
adaptive class-weighted focal loss addressing class
imbalance. Evaluated on the BUSI dataset under five-fold
stratified cross-validation and a held-out test set,
HADS-Net achieves 96.58\% test accuracy, macro
ROC-AUC of 0.9978, and macro F1-score of 0.9654,
surpassing state-of-the-art methods. Critically, no
malignant lesion is classified as normal. Future work
will focus on multi-institutional validation, learnable
edge detection, and attention-based explainability to
support clinical deployment.


\end{document}